\documentclass[11pt, twocolumn]{article}
\usepackage{aaai}
\usepackage{tikz}
\usetikzlibrary{calc,patterns,angles,quotes}
\usepackage{pdfpages}
\usepackage{times}
\usepackage{url}
\usepackage{latexsym}
\usepackage{pbox}
\usepackage{amssymb}
\usepackage{dirtytalk}
\usepackage{graphicx}
\usepackage{float}
\usepackage{pstricks}
\usepackage{pst-node}
\usepackage{pst-tree}
\usepackage{amsmath}
\usepackage{amsfonts}
\usepackage{multirow}
\usepackage{subcaption}

{\obeyspaces\gdef {\ }}
\global\newbox\codebox
\global\newbox\savedcodebox
\gdef\sverbatim{\bgroup\def\endsverbatim{\egroup\egroup\egroup\mbox{\box\codebo\
x}}\def\savecode{\egroup\egroup\egroup\global\setbox\savedcodebox\copy\codebox}\
\def\par{\egroup\vspace{-0.3em}\hbox\bgroup}\tt\obeylines\obeyspaces\global\set\
box\codebox\vbox\bgroup\hbox\bgroup}

\title{Learning event representation: As sparse as possible, but not sparser}

\author{Tuan Do and James Pustejovsky\\
  Department of Computer Science\\ Brandeis University
\\
Waltham, MA 02453 USA \\
  {\tt \{tuandn,jamesp\}@brandeis.edu}  
}

\date{}

\begin{document}
\maketitle
\begin{abstract}

Selecting an optimal event representation is essential for event classification in real world contexts. In this paper, we investigate the application of qualitative spatial reasoning (QSR) frameworks for classification of human-object interaction in three dimensional space, in comparison with the use of quantitative feature extraction approaches for the same purpose. In particular, we modify QSRLib \cite{gatsoulis2016qsrlib}, a library that allows computation of Qualitative Spatial Relations and Calculi, and employ it for feature extraction, before inputting features into our neural network models. Using an experimental setup involving motion captures of human-object interaction as three dimensional inputs, we observe that the use of qualitative spatial features significantly improves the performance of our machine learning algorithm against our baseline, while quantitative features of similar kinds fail to deliver similar improvement. We also observe that sequential representations of QSR features yield the best classification performance.  A result of our learning method is a simple approach to the qualitative representation of 3D activities as compositions of 2D actions that can be visualized and learned using 2-dimensional QSR.

\end{abstract}

\section{Introduction}
\label{intro}

The study of events has long been a focus of many disciplines, including philosophy, cognitive psychology, linguistics, computer science, and AI. \cite{tulving1983elements} postulated a separate cognitive process for event recognition called \textit{episodic memory}. In natural language, events have been studied from many different approaches, from formal logic and AI \cite{Allen:1984}, to computational linguistics \cite{TimeML-LRE:2011}. In computer science, event representations has been acquired by means of classification  \cite{shahroudy2016ntu} or represented as the composition of primitive actions \cite{veeraraghavan2007learning}.

In this paper, our focus lies within  a smaller and more restricted set of human activities involving human and object interactions. We use a fine-grained event capture and annotation framework \cite{do2017esann} as our basic setup for event classification. This framework takes into account both subeventual structures and  \textit{extra-verbal factors} \cite{Pustejovsky1995} in treating the classification of events (considering the difference between actions such as "jump on" and "jump over"). This allows us to investigate the effects of QSR on our event classification framework.

In particular, we use an event capture and annotation tool called ECAT \cite{do2016ECAT}, which employs Microsoft Kinect\circledR\ to capture sessions of performers interacting with two types of objects in our Block World environment, a cube (which can be slid on a flat surface) and a cylinder (which can be rolled). Objects are tracked using markers fixed to their sides facing the camera. They are then projected into three dimensional space using Depth of Field (DoF). Performers are tracked using the Kinect\circledR API, which provides three dimensional inputs of their joint points (e.g., wrist, palm, shoulder). Event sessions are first sliced, and each slice is then annotated with a textual description with our event language. The event descriptions are in turn parsed into tuples of semantic roles. 

We then transform the raw data into a spectrum of feature types. The first type consists of quantitative features reflecting positions of humans and objects projected on 3-dimensional and 2-dimensional feature spaces. The second type is built on top of the first, producing qualitative spatial (QS) representations for each image frame. The third type is a QS representation for each whole event duration, summarized on top of second-type features. Subsequently, we then compare our ML methods on 7 different kinds of features. 

Depending on the form of extracted features (sequence vs. non-sequence), two different machine learning methods are implemented. For frame-sequence learning, we use Long-short term memory (LSTM); for whole-event learning, we used Multilayer perceptron model (MLP). Their similar neural network structures allow us to compare  the two methods in a fair manner. In addition, we add a layer of constraints using a Conditional Random Field (CRF) algorithm before generating outputs.

The main contributions of our study are twofold. First, we propose a framework for event recording, annotation and classification, that achieves high accuracy using qualitative spatial reasoning for feature extraction. Second, by analyzing different levels of feature representations, from dense and continuous to sequential and discretized  to summary event-level features, we determine the most economical and effective way for classification of human-object interaction. 

\section{Related Work}

Human activities such as \textit{running}, \textit{sitting}, \textit{eating}, and  \textit{playing sport} have been investigated in previous research, such as  \cite{shahroudy2016ntu} and  \cite{dubba2015learning}. These human activities have significantly different motion signatures and durations. In some cases, algorithms learning to distinguish these events are actually learning the distinction  between background or color histograms. Recently, some studies have begun to introduce datasets with more complex activities, especially involving human-object interactions, such as cooking activities \cite{rohrbach2012database}, or human-human interactions, such as hand shaking \cite{ryoo2010ut}. More recently, \cite{li2016recognition} investigates the possibility of predicting partial activities using a hierarchy label space. These studies have gradually led to a more fine-grained treatment of event classification.

To facilitate event classification, it is necessary to present events  in a learnable format. This introduces the question of how to represent events, namely the difficulty in defining their temporal and spatial extensions, as well as the difficulty in selecting an observational perspective. For example, it is generally hard to demarcate an event duration from other events building up to it or  its consequences. \textit{He kicks the ball} may or may not involve the person running up to the ball, or include the ball flying to the goal. Similarly, to pick out a set of objects to include in an event representation is not trivial, even for events described in a text. For example, in \textit{He fried an egg}, should we include the frying pan in the event representation or not? Point of view (POV) is also typically under-specified, even though there can be an infinite number of ways to interpret an event, depending on the rendering location. This  leads to many different approaches in the representation of events for classification in computer science and machine learning. Events can be represented atomically, i.e., entire events are predicted in a classification manner \cite{shahroudy2016ntu}, or as combinations of more primitive actions \cite{veeraraghavan2007learning}, i.e., complex event types are learned based on recognition of combined primitive actions. For the former type of event representation, there are quantitative approaches based on low-level pixel features such as in \cite{le2011learning} and qualitative approaches such as induction from relational states among event participants \cite{dubba2015learning}. For the latter approach, systems such as \cite{hoogs2008video}, use state transition graphical models such as Dynamic Bayesian Networks (DBN). 

Event classification using qualitative spatial methods has been discussed  a fair amount in the literature. \cite{suchan2013perceptual} use the Regional Connected Calculus (RCC5) adjusted for depth field with data also recorded by Kinect\circledR sensor. This work classifies events related to people moving, sitting, or passing each other. \cite{dubba2015learning} provide an interesting framework which gives a summary explanation for a sequence of observations by alternating between  inductive and abductive commonsense reasoning. This work was applied for two activity types, one being activities happening at an airport as boundary boxes of aircraft and trucks being tracked, and one of humans interacting with an object and with each other. 

\section{QSRLib and extension}

For our study, we use the following feature types from QSRLib:

\begin{itemize}
    \item {\sc Cardinal direction}, \cite{andrew1991qualitative} (QSRLIb \textit{cardir}) measures compass relations between two objects into canonical directions such as North, North East etc. 
    \item {\sc Moving} or {\sc static} (QSRLIb \textit{mos}) measures whether a point is moving or not.
    \item {\sc Qualitative Distance Calculus}   (QSRLib \textit{argd}) discretizes the distance between two moving points. There is a significant  literature supporting the use of discretization for feature embedding. \cite{yang2009discretization} shows that \say{discretization is equivalent to using the true probability density function}. More recently, \cite{xiangjiang2017} has used this method for classification of GPS trajectory. They studied three different approaches for   discretization, including \textit{equal-width binning}, Recursive Minimal Entropy
    Partitioning (RMEP) \cite{dougherty1995supervised} and fuzzy discretization\cite{roy2003fuzzy}. Their finding is that the \textit{equal-width binning} approach is both simple and effective, so we used this approach with an interval length of 1/20 meter in embedded space. We did not  explore other interval lengths, leaving that for future experiments.
    \item {\sc Qualitative Trajectory Calculus} (Double Cross) (QSRLib \textit{qtccs}): $QTC_C$ is a representation of motions between two objects by considering them as two moving point objects (MPOs) \cite{delafontaine2011implementing}. The type C21 of $QTC_C$ (implemented in QSRLib) considers whether two points are moving toward each other or whether they are moving to the left or to the right of each other. Apparently, this is the kind of spatial semantics needed to learn the prepositions we used in this experiment. The following diagram explains this:

\begin{tikzpicture}
    \coordinate (k) at (1, 0);
    \coordinate (l) at (5, 0);
    \coordinate (v) at (2, 1);
    \coordinate (u) at (2.5, 0.5);
    \fill[black] (k) circle (0.1) node[black,anchor=north east] {$k$};
    \fill[black] (l) circle (0.1)
    node[black,anchor=north east] {$l$};
    
    \draw[thick,black,-] (k) -- (l) ;
    \draw[thick,gray,-] (k) -- (v) ;
    \draw[thick,gray,-] (k) -- (u) ;
    \pic [draw, ->, "$\alpha$", angle eccentricity=1.5, angle radius = 0.5cm] {angle = l--k--v};
    \pic [draw, ->, "$\beta$", angle eccentricity=1.5, angle radius = 1.0cm] {angle = l--k--u};
     
    \draw[loosely dotted,black,-] (k) -- ++(0,-1.5);
    \draw[loosely dotted,black,-] (k) -- ++(0,1.5);
    \draw[loosely dotted,black,-] (k) -- ++(-1.5,0);
    \draw[loosely dotted,black,-] (l) -- ++(0,-1.5);
    \draw[loosely dotted,black,-] (l) -- ++(0,1.5);
    \draw[loosely dotted,black,-] (l) -- ++(1.5,0);
\end{tikzpicture}

\begin{figure*}
\centering
\noindent\includegraphics[width=1\textwidth]{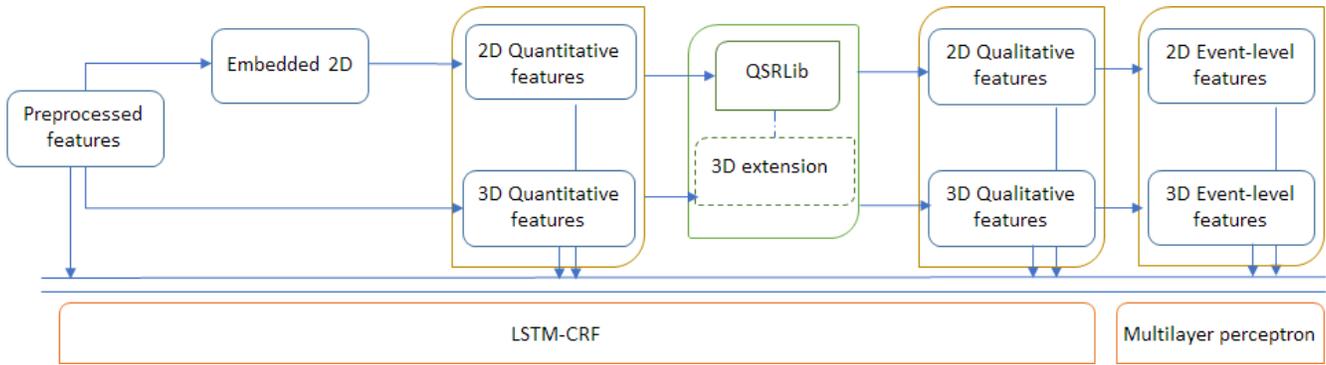}
\vspace{8mm}
\caption{Downstream feature extraction methods used in this study. Our focus is on the performance gain from quantitative features to qualitative features through the use of QSRLib and its extension for 3-dimensional data}
\label{fig:feature-extraction}
\end{figure*}

$QTC_C$ produces a tuple of 4 slots $(A, B, C, D)$, where each could be given either -, + or 0, depending on the angle $\alpha$. For example, C is $+$ if $\alpha > 0 \land \alpha < 180$, $-$ if $\alpha > 180 \land \alpha < 360$ and 0 otherwise. QSRLib also allows specification of a \textit{quantisation factor} $\theta$, which dictates whether the movement of a point is significant in comparison to the distance between $k$ and $l$.
\end{itemize}

\begin{figure}[ht]
    \centering
    \includegraphics[width=0.4\textwidth]{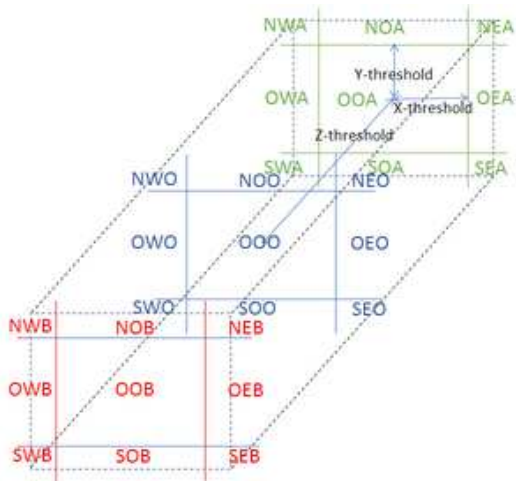}
    \vspace{6mm}
    \caption{3D Grid around a point. N, S, W, E, B, A stand for North, South, West, East, Below, Above.}
    \label{fig:cardinal-direction}
\end{figure}

\textbf{Modification of Qualitative trajectory calculus Double Cross ($QTC_C$) implementation in QSRLib}: 

 $QTC_C$ is not the most appropriate calculus for use in event classification without some modification, since approximating objects as MPOs leads to loss of representational information. Exact modeling of object 3D volumes are feasible but cumbersome. 

\begin{figure}[ht]
    \centering
    \includegraphics[width=0.2\textwidth]{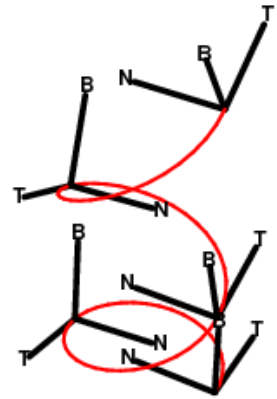}
    \vspace{6mm}
    \caption{Frenet-Serret frame for 3-D qualitative trajectory calculus}
    \label{fig:frenet-serret}
\end{figure}

Our solution for this involves an \textit{angle quantisation factor} $\beta > 0$ which should be relatively small. When $|\alpha| < \beta$ or $|\alpha - \pi| < \beta$, we set the the value of C to 0. Similarly for slot A, when $|\alpha - \frac{\pi}{2}| < \beta$ or $|\alpha - \frac{3\pi}{2}| < \beta$, we also set the value to be 0. 

\textbf{3-dimensional extension} We extended QSRLib for the following feature types:

\begin{itemize}
    \item {\sc Cardinal direction (3D)}: We followed the 3D grid approach to partition the space into 3x3x3=27 voxels, \cite{sabharwal2014modeling}. The center voxel is the Minimum Bounding Hyper-rectangle (MBHR) of a reference object. In this framework, given two objects A (reference object) and B (target object), the cardinal direction from A to B is calculated firstly by generating the MBHR of A (it can be approximated), then setting up the 3D grid for A, then finding which voxels intersect with B. We actually used a much simpler alternative that replaces B with its centroid, so that only one direction is resulted.  

    \item {\sc Qualitative Distance Calculus (3D)}: This is basically the same as for two dimensional.
    
    \item {\sc Qualitative Trajectory Calculus (3D)}: It is noted that there are no features analogous to lateral slots $(C,D)$ for 3-dimension. \cite{mavridis2015qtc} provides an alternative for lateral relations, using Frenet-Serret frames (FS frame). We do not give the details of the calculation here, but rather provide our implementation. In a nutshell, the calculation steps are as follows:
    \begin{itemize}
        \item For each point P and Q, calculate the tangent vector, the binormal vector and the normal vector. For continuous data domain, this requires calculating second derivative of each point's moving curve. For discrete domain, this translates to taking three data points into account for each calculation, of the current and two previous time steps. These three vectors create a FS frame for each moving point. 
        
        \item Assuming that two FS frames $F_P$ and $F_Q$ are calculated (special values are assigned in the case of degeneration), we need to find a transformation matrix from $F_P$ to $F_Q$. This rotation matrix is in turn decomposed into three values of yaw, pitch and roll angles. These three values, together with two feature values $(A,B)$ in $QTC_C$ make a tuple of 5 values in our $QTC_{3D}$.
        
    \end{itemize}
\end{itemize}

\section{Feature extraction}

Figure~\ref{fig:feature-extraction} shows our downstream feature extraction methods.

Our motivation for creating downstream features is of the following basic intuition.
\begin{itemize}
    \item Object Model: State-by-state characterization of an object as it changes or moves through time.
    \item Action Model: State-by-state characterization of an actor’s motion through time. When action involves multiple objects, this also includes effect of objects on each other. 
    \item Event Model: Composition of the object model with the action model.
\end{itemize}

From a recognition point of view, the object model is translated to inherent motion of objects whereas the action model is translated to inter-object relative motion. 
\begin{align*}
M_E = M_R * \prod_{i\in[1,2]}{M_{Oi}} * M_{O1O2} * \prod_{i\in[1,2]}{M_{ROi}}
\end{align*}
Here,  $R$ stands for human body rig, and $O_i$ stands for objects. This method factorizes the model representation into $n * (n + 3) / 2$ terms where $n$ is the number of objects. This is not a very economical  representation when there is a large number of objects in the scene ($O(n^2)$), but in reality, the  number of objects is relatively static to each other, or we in fact need to consider a smaller number of objects to allow for possible descriptions (think of ``Lunar eclipse occurs when the Moon passes directly behind the Earth into Earth's shadow, aligning with the sun and Earth'', where we in fact do not take into account movement of the other planets). 

\subsection{Preprocessing}
The raw data come from two sensors on Kinect\circledR: the RGB camera and the time of flight (ToF) depth sensor. In turn, these produce three streams of vision input: a stream for RGB, a stream for depth field, and a stream for tracked human body rigs. These data streams have different rates and resolutions.

\textbf{Tracking object}: We used the Glyph detection algorithm \cite{glyph}, with some adjustments to detect Glyph markers stuck on the objects (Glyph markers are black and white checked square). Markers put on different objects are distinguished to simplify the tracking process. For frames where tracking is lost, the marker's 2D position is interpolated. 2-dimensional data is projected into 3-dimensional by using depth field. Body rig joint points are already tracked by Kinect\circledR's SDK.

\textbf{Normalizing rate}: Different streams of data were regenerated with the same rate by re-sampling with interpolation. We used the rate of 24fps, which is the same as the RGB stream.  

\subsection{Quantitative features}

\textbf{3D features} are generated by the following methods:

\begin{itemize}
\item Relative motions between different objects are approximated by calculating the distance vectors among these entities. 
\item To model human body rigs, vectors among the following points are calculated: middle point between the shoulders, left hand tip and right hand tip. 
\item To model objects, vectors between two diagonally opposing points are used.
\end{itemize}

\textbf{Embedded 2D features}

For each factor model, we used Principle Component Analysis (PCA) to project points considered into 2-dimensional planes, with the hope that the kept dimensions will keep the maximum variation, while provide an efficient way to visualize and reason about the data.  The set of features for each frame is analogous to the 3D case, but with all data points from each factor model projected through PCA.

\subsection{Frame-level qualitative features}

\begin{figure}[h]
    \centering
    \begin{subfigure}{.5\linewidth}
      \centering
      \includegraphics[width=1.0\linewidth]{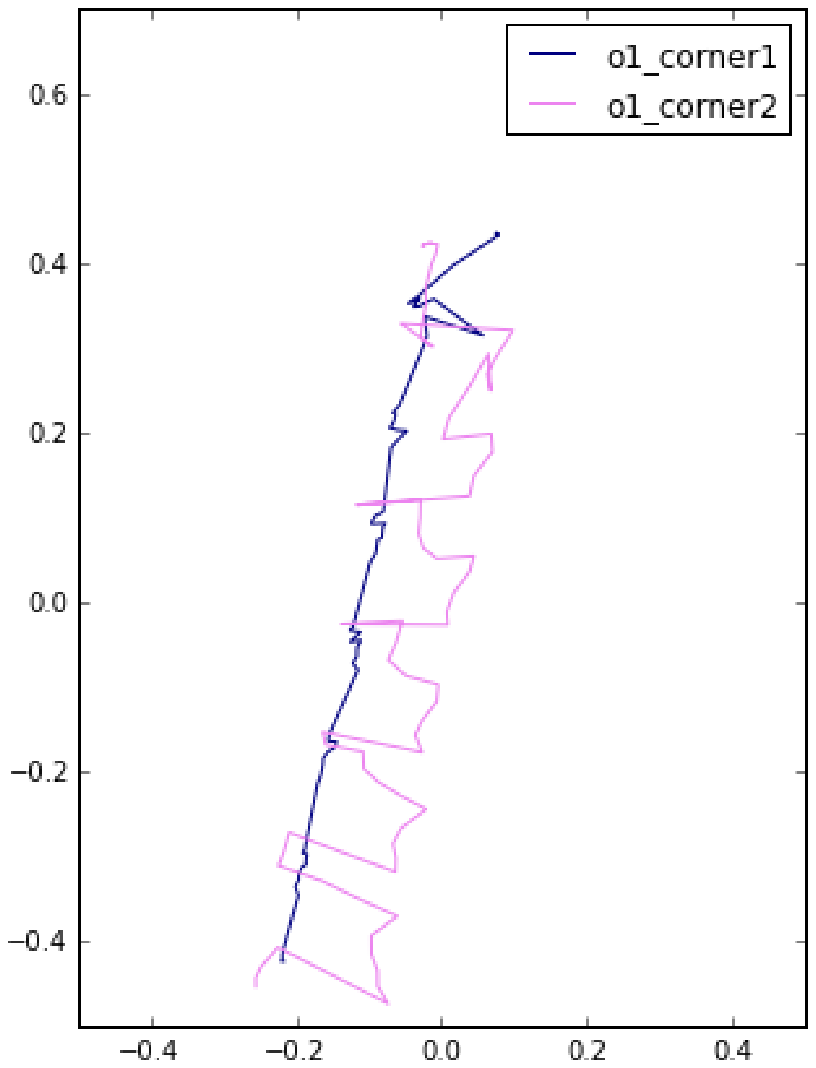}
      \caption{Corners of a rolling object}
      \label{fig:corners}
    \end{subfigure}%
    \begin{subfigure}{.5\linewidth}
      \centering
      \includegraphics[width=1.0\linewidth]{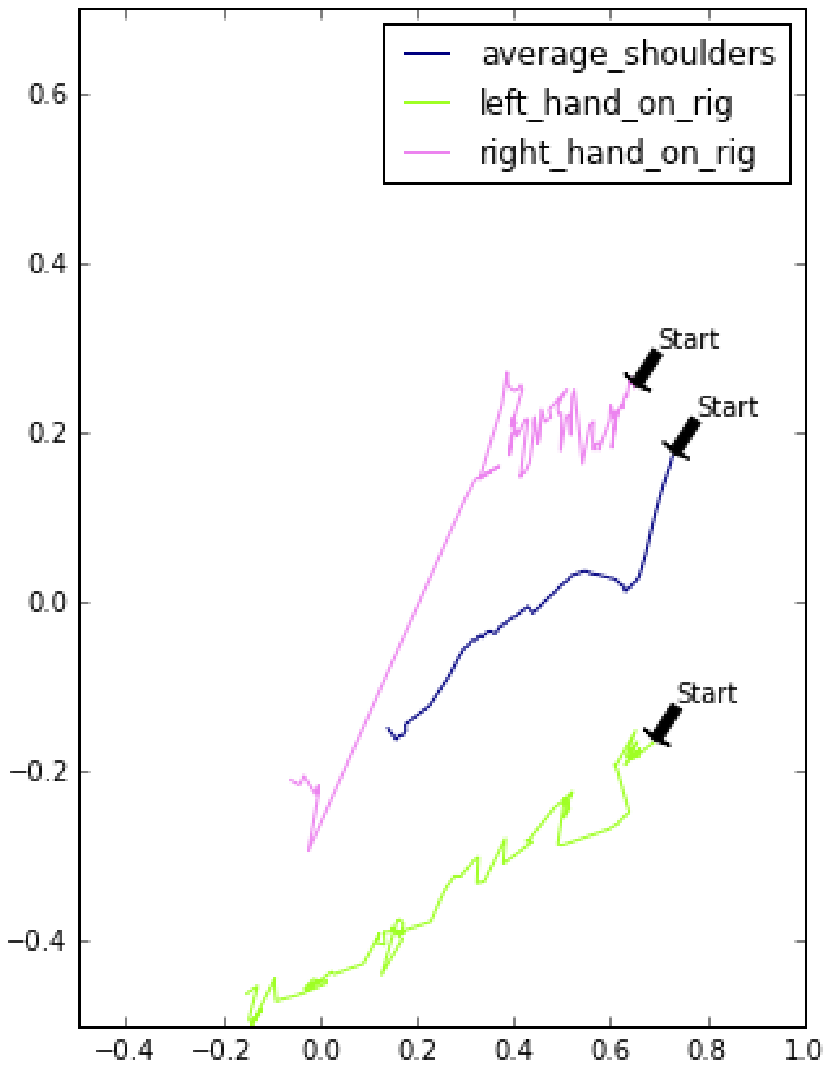}
      \caption{Hands in relative to body}
      \label{fig:hands}
    \end{subfigure}
    \vspace{6mm}
    \caption{Samples of trajectories in embedded spaces}
    \label{fig:trajectories}
\end{figure}

The set of qualitative features are \textit{downstream} features from the set of quantitative features. We employed 4 feature types as listed before. Visualizing trajectories of objects in embedded spaces gave us insight into the quality of our extracted features. For example, we observed that 2D qualitative trajectory calculus has a strong explanatory power in distinction of inherent motion of objects. For example, Figure~\ref{fig:corners} shows typical trajectories of two corner points of a rolling object. Direction between these points shows periodical change that is easily captured as a change of cardinal directions in the feature space.

\subsection{Event-level qualitative features}

A \textit{downstream} set of event features would require a method to summarize the change of frame-level features across the event duration. There are multiple ways to do that, one is by specifying a set of primitive actions, such as in \cite{suchan2013perceptual}. This approach requires a hierarchical \textit{fuzzy decomposition} of event into subevents. 

Another approach that employs a form of Inductive Logic Programming (ILP) is that of \cite{dubba2015learning}. This approach is based on Inductive-Abductive reasoning framework, in which the authors look for \textit{basic, minimal, constraint satisfied} narratives that explain changes of RCC-states as observed from event captures for each event type. This framework is quite interesting and might be effective, especially when event interpretation depends only on change of RCC-states (\textit{dc, touch, in}). We, however, are not sure how to adapt this framework for other spatial qualitative relationships, because that would requires generation of intermediate qualitative states for each object pair. For an \textit{ordered} set of states in RCC, it might be feasible, but for cardinal directions or QTC states, the path between two states is not unique, and the number of intermediate states to be considered would increase quickly. For a small set of training data, that might lead to a form of overfitting, as too many rules are produced.

For this reason, we resort to use a simple and \textit{feature-based only} representation. In fact, we use the frame-level features of the first frame and the last frame, plus the different vector between these two. 

\section{Learning algorithms}

\subsection{Multilayer perceptron learning (MLP)}

For event-level features, we use a simple multilayer perceptron model for comparison with our sequential neural network models. The multilayer perceptron (MLP) is one of the earliest neural network model, employing a feedforward infrastructure with backpropagation update. Here we use a rectified MLP (a stack of layers in which each has a linear layer combined with a Rectified Linear Unit \cite{glorot2011deep}). Dropout is also applied to reduce overfitting.

\subsection{Long-short term memory (LSTM)}

LSTM is a flavor of deep Recursive neural network (RNN) that has generally solved the problem of \say{vanishing gradients} in traditional RNN learning \cite{hochreiter1997long,schmidhuber2015deep} and has found their application in a wide range of problems involving sequential learning, such as hand written recognition, speech recognition, gesture recognition, etc. 

\begin{figure}[!htbp]
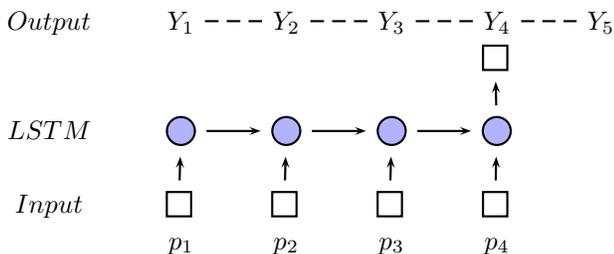

\begin{center}
$
\psmatrix[colsep=1cm,rowsep=0.1cm]
Output & [name=Y_1] Y_1 & [name=Y_2] Y_2 & [name=Y_3] Y_3 & [name=Y_4] Y_4 & [name=Y_5] Y_5 \\
& & & & [name=O, mnode=f]\\
 \\
LSTM & [name=H_1, mnode=C,radius=0.2,fillstyle=solid,fillcolor=blue!30] & [name=H_2, mnode=C,radius=0.2,fillstyle=solid,fillcolor=blue!30] & [name=H_3, mnode=C,radius=0.2,fillstyle=solid,fillcolor=blue!30] & [name=H_4, mnode=C,radius=0.2,fillstyle=solid,fillcolor=blue!30] \\
\\
Input & [name=I_1,mnode=f]  & [name=I_2,mnode=f] & [name=I_3,mnode=f] & [name=I_4,mnode=f]  \\
& p_1 & p_2 & p_3 & p_4
\ncline[nodesep=4pt,linestyle=dashed]{-}{Y_4}{Y_5}
\ncline[nodesep=4pt,linestyle=dashed]{-}{Y_3}{Y_4}
\ncline[nodesep=4pt,linestyle=dashed]{-}{Y_2}{Y_3}
\ncline[nodesep=4pt,linestyle=dashed]{-}{Y_1}{Y_2}
\ncline[nodesep=4pt]{->}{H_4}{O}
\ncline[nodesep=4pt]{->}{H_3}{H_4}
\ncline[nodesep=4pt]{->}{H_2}{H_3}
\ncline[nodesep=4pt]{->}{H_1}{H_2}
\ncline[nodesep=4pt]{->}{I_1}{H_1}
\ncline[nodesep=4pt]{->}{I_2}{H_2}
\ncline[nodesep=4pt]{->}{I_3}{H_3}
\ncline[nodesep=4pt]{->}{I_4}{H_4}
\endpsmatrix
$
\end{center}
\hspace*{-2cm}
\caption{LSTM model with possible constraints of outputs with CRF. CRF layer is represented as dashed links among predicted labels. }
\end{figure}

We will not describe in detail the LSTM model, as it has become a standard sequential learning method. Interested readers are recommended to take a look at either our implementation \footnote{https://github.com/tuandnvn/ecat\_learning} or a popular reference on RNN and LSTM \footnote{http://www.wildml.com/2015/10/recurrent-neural-network-tutorial-part-4-implementing-a-grulstm-rnn-with-python-and-theano/}. Briefly, however, the model passes each feature vector through a linear layer before feeding each sequence into an LSTM. Each label $Y_i$ requires a separate LSTM cell, $X_i$. Output of each LSTM cell is a term $t, t \in {t_s, t_v, t_o,  t_p, t_l}$ corresponding to 5 semantic slots in the tuple. We will combine these values with our CRF weights, discussed in LSTM-CRF.

\subsection{Conditional Random Field (CRF)}

\newcommand{\shb}[1]{[name=#1,mnode=r] \psframebox[shadow=true, framearc = 0.25,framesep=0.03]{#1}}

\begin{figure}[!htbp]
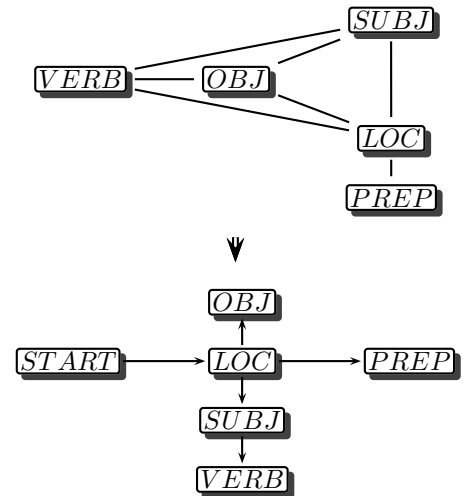

\centering
$
\psmatrix[rowsep=1.0cm]
[name=original-CRF]
\psmatrix[colsep=1cm,rowsep=0.4cm]
&&  \shb{SUBJ} \\
\shb{VERB} & \shb{OBJ} \\
&& \shb{LOC}\\
&& \shb{PREP}
\ncline[nodesep=3pt]{SUBJ}{OBJ}
\ncline[nodesep=3pt]{OBJ}{LOC}
\ncline[nodesep=3pt]{SUBJ}{LOC}
\ncline[nodesep=3pt]{SUBJ}{VERB}
\ncline[nodesep=3pt]{OBJ}{VERB}
\ncline[nodesep=3pt]{LOC}{VERB}
\ncline[nodesep=3pt]{LOC}{PREP}
\endpsmatrix
\\
[name=tree-CRF]
\psmatrix[colsep=1cm,rowsep=0.4cm]
& \shb{OBJ} \\
\shb{START} & \shb{LOC} & \shb{PREP}  \\
& \shb{SUBJ} \\
& \shb{VERB}
\ncline[nodesep=1pt]{->}{START}{LOC}
\ncline[nodesep=1pt]{->}{LOC}{SUBJ}
\ncline[nodesep=1pt]{->}{LOC}{OBJ}
\ncline[nodesep=1pt]{->}{LOC}{PREP}
\ncline[nodesep=1pt]{->}{SUBJ}{VERB}
\endpsmatrix
\ncline[doubleline=true,nodesep=10pt]{->}{original-CRF}{tree-CRF}
\endpsmatrix
$
\vspace*{0.5cm}
\caption{Reformation from general {\bf CRF} (left) to {\bf Tree-CRF} (right)}
\label{fig:crf}
\end{figure}

CRF has been used extensively to learn structured output as it allows specification of constraints of output labels \cite{sutton2006introduction}. In this model we wish to constrain the outputs so that: one object (Performer or the other objects) is not allowed to fill  two different syntactic slots; when there is no verb, all the other slots should be None; locative and preposition are dependent, because if locative is None, preposition must also be None and vice versa. The edges between nodes on the left side of Figure~\ref{fig:crf} show the dependencies on output labels that we wish to model.

However, training and classifying using a full CRF model would be more difficult, especially when implemented  with a neural network architecture.  We modified the model into a tree-CRF structure (right side of Figure~\ref{fig:crf}) to make the model learnable using dynamic programming. The complexity of the algorithm reduced from $O(n^5)$ to $O(n^2) * 5)$ where $n$ is the size of our vocabulary. The learning problem is thereby changed to learning the weights along the edges on the tree-CRF, for example, $P\_locative\_preposition$ (together with parameters of LSTM). The directionality of edges is the forward direction of the message passing algorithm used for learning (and in reverse, for testing using the backward direction).

\subsubsection{Use of CRF as a constraint layer}

Naturally, as we have learned 5 models (either 5 MLP or 5 LSTM) to predict 5 values in the output, we want to regularize the output combination by CRF. Moreover, LSTM-CRF is a natural extension of LSTM applied for constrained outputs. For instance it is used for named entities recognition task to model constraints on BIO labels \cite{huang2015bidirectional}. 

To put CRF learning on top of MLP or LSTM, we modify the term $t$ (the term before softmax) produced by outputs of MLP or LSTM as followings.

\begin{align*}
t(l,s,o,p,v) &= t_l + t_s + t_o + t_p + t_v \text{\vspace{20mm} \textbf{     Original target}} \\
t(l,s,o,p,v) &= t_l + t_s + t_o + t_p + t_v \text{\vspace{20mm} \textbf{     Modified target}}\\
&+ P_{start\_l} + P_{ls} + P_{lo} + P_{lp} + P_{sv}
\end{align*}

\noindent where $l$, $s$, $o$, $p$, $v$ stand for Locative, Subject, Object, Preposition and Verb respectively. 

In training, $softmax$ is calculated for a predicted label combination, namely $(l', s', o', p', v')$ as below. We can calculate the $log$ of $sum$ using message passing over the tree nodes of the CRF tree. We use cross entropy between predicted distribution and correct output as the $cost$ in training.

\begin{align*}
so&ftmax = exp[ t (l',s',o',p',v') - \\
& log [\sum_{l}\sum_{s}\sum_{o}\sum_{p}\sum_{v} exp(t (l,s,o,p,v)) ] ] \\
&=  exp[ t (l',s',o',p',v')
-\\
& log [ \sum_{l} exp(t_l + P_{start\_l})
[\sum_{s} exp(t_s + P_{ls}) \\ 
& \sum_{v} exp(t_v + P_{sv})]
[\sum_{o} exp(t_o + P_{lo})] [\sum_{p} exp(t_p + P_{lp})]]
\end{align*}

\noindent In evaluation, a similar message passing algorithm is used, but instead of $log\_sum$, we use $max$ to calculate the probabilities and $argmax$ to keep track of the best combination.

\section{Experimental setup}

\begin{figure}[h]
    \centering
    \includegraphics[width=0.4\textwidth]{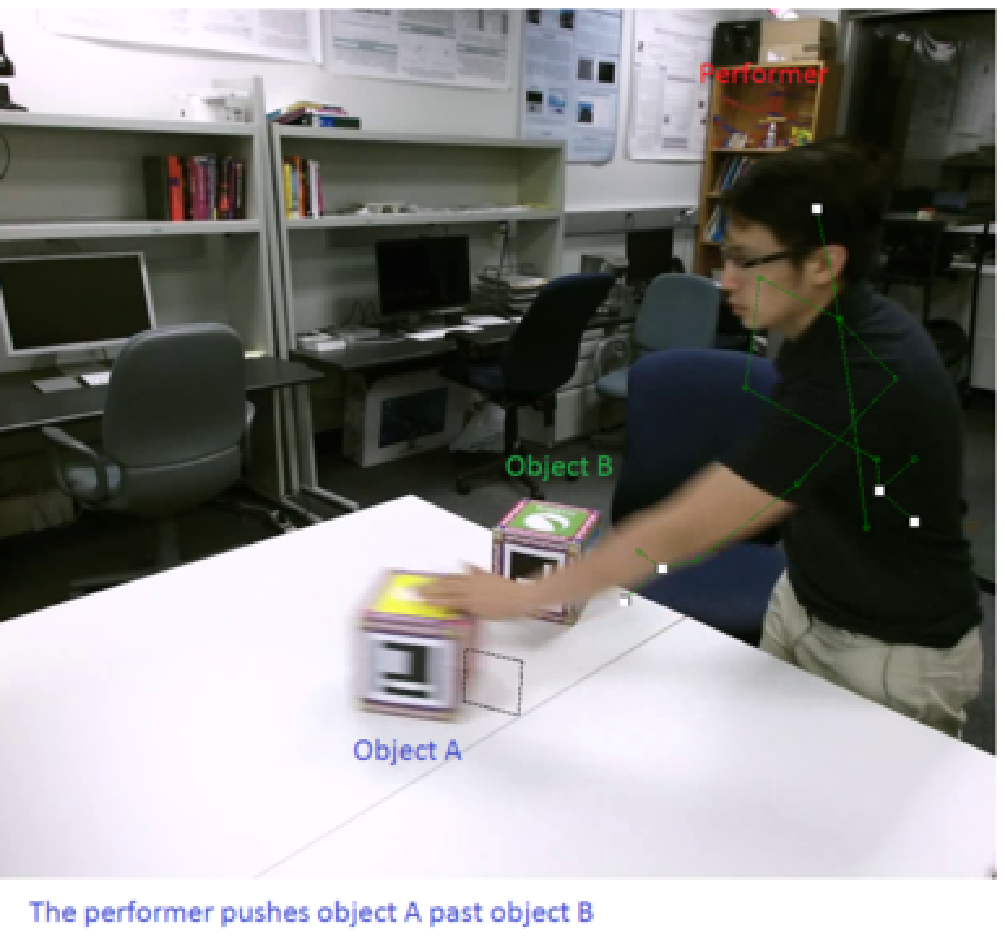}
    \vspace{6mm}
    \caption{Event capture with fine-grained annotation}
    \label{fig:event-capture}
\end{figure}

To demonstrate our model's capability to learn the \textit{spatio-temporal} dynamics of object interactions in events, we use a collection of four action types: \textit{push}, \textit{pull}, \textit{slide}, and \textit{roll}, along with  three different \textit{spatial} prepositions used for space configurations between objects, namely \textit{toward} (when the trajectory of a moving object is straightly lined up with a destination static object and makes it closer to that target), \textit{away from} (makes it further from that object) and \textit{past} (moving object getting closer to static object then further again).

Afterwards, for each session, we sliced the events into short segments of 20 frames. Two annotators were assigned to watch and annotate them (segments can be played back). To speed up annotation, only event types related to original captured types are shown for selection. For instance, if  the event type of the captured session is \say{The performer pushes A toward B}, other available event types are \say{The performer pushes A}, \say{A slides toward B} or \say{None}.

We explored different combinations of hyperparameters for our MLP and LSTM models. Two methods are used to combat over-fitting: (i) dropout \cite{hinton2012improving} (for LSTM this a dropout wrapper on LSTM cell, for MLP this is a dropout wrapper on each layer) and (ii) gradient clipping by a global norm. Following is the list of hyperparameters we used for tuning:
\begin{itemize}
    \item Number of LSTM layers: [1, 2]
    \item Size of hidden layer: [200, 400]
    \item Learning rate: [0.05, 0.1, 0.2, 0.5]
    \item Dropout rate: [0.5, 0.6, 0.8]
    \item Learning rate decay: [0.94, 0.95, 0.96]
\end{itemize}
The network is trained with mini-batch gradient descent optimization for 200 epochs (LSTM) or 500 epochs (MLP) on the Tensorflow library. 

\section{Results}

Captured sessions are split for 5-fold cross validation, i.e., 24 sessions for training and 6 for testing on each fold. We use the LSTM-CRF model with raw input data as the baseline. A prediction is correct if all slots are correct. Performance is reported in the following tables:

\begin{table}[!ht]
\scriptsize
\label{table:evaluation}
\centering
\begin{tabular}{|l||l|c|}
\hline
& Model & Precision \\
\hline
\multirow{4}{*}{Frame level} & 3D-Raw-LSTM-CRF & 43\% \\
\cline{2-3}
& 3D-Quant-LSTM-CRF & 44\% \\
\cline{2-3}
& 3D-Qual-LSTM-CRF & 52\% \\
\cline{2-3}
& 2D-Quant-LSTM-CRF & 48\% \\
\cline{2-3}
& 2D-Qual-LSTM-CRF & \textbf{60\%} \\
\hline
\multirow{2}{*}{Event level} & 3D-Event-Qual-LSTM-CRF & 20\% \\
\cline{2-3}
& 2D-Event-Qual-LSTM-CRF & 23\% \\
\hline
\end{tabular}
\vspace{6mm}
\caption{Evaluation}
\end{table}
\begin{table}[!ht]

\scriptsize
\label{table:best}
\centering
\begin{tabular}{|l|c|}
\hline
Label & Precision \\
\hline
Subject & 93\% \\
\hline
Object & 90\% \\
\hline
Locative & 80\% \\
\hline
Verb & 83\% \\
\hline
Preposition & 82\% \\
\hline
\end{tabular}
\vspace{6mm}
\caption{Label precision breakdown for Frame-2D-Qual-LSTM-CRF. Note that because the labels are not independent, precision of the joint model is not the product of individual figures}
\end{table}

\noindent We can observe a significant improvement of classification using 2-dimensional frame-level qualitative features. Frame-level quantitative features did improve over our baseline, but the improvement is not as impressive. Moreover, summarizing frame-level features to create event-level features creates a lossy representation that is not be able to learned efficiently.

Given these results, it is worth considering possible explanations for our findings. Firstly, as pointed out by \cite{yang2009discretization} and \cite{xiangjiang2017}, qualitative representation is a method of discretization, which makes data sparser, therefore easier to learn. Especially when taking the difference between features of two adjacent frames, as a qualitative feature strongly distinguishes between 0 and 1, the effect of sequential change is more pronounced. 

Moreover, we observed that the best performance for \textbf{Qual-LSTM-CRF} is achieved by configuring two layers of LSTM with 400 nodes on the hidden layers, while for other models, the number of layers does not affect the performance significantly. Differently from a feed-forward neural network such as Convolutional Neural Network (CNN), which can learn more abstract and useful features when it get deeper, LSTM needs some help from the representation of features to reap a benefit from going deeper. 


We also learned that a simple summary representation for events is not effective. We were very surprised that the results from event level features were much lower than we expected. Though we did expect that the loss of information regarding path contour could make the classifier indifferent to the distinction between \text{rolling} and \text{sliding}, that alone might not be able to explain the bad performance. Another possible factor is that there are many frames where the tracking is either lost or very noisy that the representation for a single frame is meaningless. If it happens to be the beginning or end frame, the representation is not of high quality. Using a sequence of frame representation has a positive effect to compensate for frame noise.

If instead we also takes into account features in intermediate frames, the representation comes back to a sequential one. There may be a better way to do this, such as summarizing for each feature separately, or using the Inductive-Abductive framework as mentioned before, but as we already discussed, it is unclear how to perform these steps efficiently. This leads us to a belief that possibly a representation, which is sparse for each point in time but dense temporally, would be reasonably easy to learn, resulting in more compact and explanatory models. 
  
\section{Conclusion and Future Directions}

In this study we have explored a wide range of feature extraction methods to learn what is the best way to represent events for classification. Even though our event domain is quite restricted or artificial, it represents complex human-object interactions, and therefore we believe that our conclusion could be generalized to other domains of similar complexity, such as that of human-robot interaction. 

We acknowledge our shortcoming in the investigation of event-level qualitative features. More complex methods, such as those in \cite{dubba2015learning} could be used in a similar manner, and we will leave that for future directions. 

One more dimension that we wanted to explore but we have not is that of the embedding method from 3-dimensional to 2-dimensional data. Methods other than PCA could be used, such as multidimensional scaling. We will also leave that for future research. 

We have started to extend this learning framework to a more generic human-object problem in which objects are of common types, such as balls, books etc, requiring a subtask of classifying objects involved in an interaction as well. With more complex variation in shape of objects in movements, we expect to deal with more difficulty in object model representation. We will base on another work that our lab has developed called VoxSim \cite{krishnaswamy2016voxsim} to scaffold an estimated representation for each object. Similar to the way that we have modeled internal movements of blocks in our Block World by qualitative change among markers' corners, movements of generic objects could be estimated by considering relative movements between their components. For example, movement of a cup could be estimated by observing the relative motion between its rim and its handle. In turn, these parts could be estimated using both visual features and the cup's \textit{Gibsonian} affordances when in interaction with performers.

Based on VoxSim simulated environment, we are also bootstrapping a dataset to be used as auxiliary training samples. The idea is that training data for our generic event classification suffer from sparseness with regards to all different positions and configurations of objects. We intend to use our environment to create a multitude of event simulations, projecting them into a same embedded space as our visual captures. We are just at the very beginning phase of doing that, and we hope that more results and discussions would be followed in the near future.

\begin{figure}[h]
    \centering
    \includegraphics[width=0.2\textwidth]{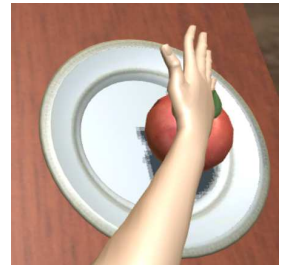}
    \vspace{6mm}
    \caption{``Put an apple on a plate'' - a generic event realization in our embedded simulated environment. From \cite{krishnaswamy2016voxsim}}
    \label{fig:visualize}
\end{figure}

\section*{Acknowledgements} 
  \label{sec:acknowledgements}
This work is supported by a contract with the US Defense Advanced Research Projects Agency (DARPA), Contract W911NF-15-C-0238.
 Approved for Public Release, Distribution Unlimited. The views expressed are those of the authors and do not reflect the official policy or position of the Department of Defense or the U.S. Government.  We would like to thank Nikhil Krishnaswamy and
Keigh Rim for their discussion and input on this topic.  All errors and mistakes are, of course, the responsibilities of the authors..
  \newpage
\bibliographystyle{aaai} 
\bibliography{References}

\end{document}